\documentclass{article}

\usepackage{neurips_2024}

\usepackage[utf8]{inputenc} 
\usepackage[T1]{fontenc}    
\usepackage{hyperref}       
\usepackage{url}            
\usepackage{booktabs}       
\usepackage{amsfonts}       
\usepackage{nicefrac}       
\usepackage{microtype}      
\usepackage{xcolor}         
\usepackage{siunitx}
\usepackage{graphicx}
\graphicspath{ {./} }

\newcommand{\beginsupplement}{%
        \setcounter{table}{0}
        \renewcommand{\thetable}{S\arabic{table}}%
        \setcounter{figure}{0}
        \renewcommand{\thefigure}{S\arabic{figure}}%
     }

\title{Need is All You Need: Homeostatic Neural Networks Adapt to Concept Shift}

\author{%
  Kingson Man \\ 
  Feeling Machines LLC \\
  \texttt{kingson@feelingmachines.net} \\
   \And
   Antonio Damasio \\
   University of Southern California\\
   \texttt{damasio@usc.edu} \\
   \AND
   Hartmut Neven \\
   Google Inc. \\
   \texttt{neven@google.com} \\
}

\begin{document}

\maketitle

\begin{abstract}
In living organisms, homeostasis is the natural regulation of internal states aimed at maintaining conditions compatible with life. Here, we introduce an artificial neural network that incorporates some homeostatic features. Its own computing substrate is placed in a needful and vulnerable relation to the very objects over which it computes. For example, MNIST digits may cause excitatory or inhibitory effects upon the homeostatic network that classifies them, by altering the network’s learning rate. Accurate recognition is desirable to the agent itself because it guides decisions to up- or down-regulate its internal states and functionality. Counter-intuitively, the addition of vulnerability to a learner can confer some benefits. Homeostatic learners are more adaptive under conditions of concept shift, in which the relationships between labels and data change over time. The greatest advantages are obtained under the highest rates of shift. Homeostatic learners are also resilient to second-order shift, or environments with changing rates of concept shift. 

\end{abstract}

\section{Introduction}

To paraphrase Heraclitus, "The only constant in life is change". The rules and relationships learned today may no longer hold tomorrow. Un-learning the bad old rules, re-learning the good new ones, and knowing how to tell the difference remains a major challenge for learning machines. Here we are inspired by the natural intelligence of living organisms, which maintain themselves in the face of environmental change by following the dictates of homeostasis. Homeostasis is the regulation of internal body states within a range compatible with life. It has been proposed that (a) machines that implement a process resembling homeostasis could be designed to exhibit a feeling-like device for the motivation and evaluation of their behavior and that (b) equipping an artificial learner with a feeling-like device might improve its adaptiveness to the inconstant data streams of the real world (Man and Damasio 2019).

Here we present a homeostatic neural network architecture in which a classifier is placed into a needful and vulnerable relation to the objects over which it computes. By way of analogy, the homeostatic agent must learn to distinguish between cups of coffee and cups of beer, while also needing to take a drink every so often to regulate its own mental arousal. In this setting, accurate classification is desirable to the agent itself because it guides decisions that can carry consequences for its internal states. 
\section{Background}
In biological brains, neurons regulate their excitability and synaptic conductance to stabilize network function (Marder and Goaillard 2006). In artificial neural networks, homeostatic regulation of excitability can reduce saturation and improve signal propagation (Williams and Noble 2007). In simulation studies of evolutionary robotics, phototactic robots used ‘neural plasticity’ to restore adaptive behavior following visual field inversion (Di Paolo 2000; Iizuka and Di Paolo 2008). However, the homeostatic-like features of prior works were implemented from the outside-in: systems were instructed to maximize, or keep within a set range, certain arbitrary values that were labeled "homeostatic". The operation of the system itself was not exposed to the consequences of the system's own activities, that is, it was not made vulnerable to the world and therefore sensitive to changes in the world.

Non-stationarity, or "changes in the world", poses a major challenge in machine learning. Learners can fail to generalize because of concept shift (Moreno-Torres et al 2012), in which the associations between labels \(y\) and observations \(x\) change across the training and testing phases: \(P_{train}(y|x) \neq P_{test}(y|x)\). This phenomenon occurs frequently in real world settings of online supervised learning; for example, recommender systems must stay current with their users' evolving tastes. 

\section{Homeostatic architecture of needful neural networks}
Our homeostatic agent learns to classify images of objects. In a twist, the learner is designed to be needful – it depends on the objects that it classifies for its continued integrity and functionality. The objects have direct effects, excitatory or inhibitory, on the learner itself. For example, in MNIST classification, the digits \{0,1,2,3,4\} have inhibitory effects and reduce the learning rate (LR), while the digits \{5,6,7,8,9\} have excitatory effects and increase the LR. (The mapping of digits to effects is arbitrary and is ordered here only for convenience.) Critically, following classification of an object, the learner decides to either "ingest" the digit and alter its own learning rate, or "reject" the digit and keep its current learning rate. We use a counterfactual decision process to answer the question, "How would my own functionality be affected by taking or leaving this object?" The learner evaluates each alternative by simulating both versions of itself and testing them against a store of recently seen objects and labels (Supplementary Fig. 1). Misperceiving an object can lead to performing the wrong simulation of the object’s effects on the learner. This will drive the wrong LR decision, further destabilizing future perceptions.

\subsection{The vicissitudes of life}
As so often happens in life, the rules have a way of changing on you. We introduce concept shift by permuting labels on a subset of the data. When a shift occurs, we swap the labels for two randomly selected classes. For the MNIST example, we may swap the labels "zero" and "nine", such that all images that look like "\(0\)" are now labelled "nine", and all images that look like "\(9\)" are now labelled "zero" (Supplementary Fig. 2). Note that in this swap, the homeostatic effects of the digits have also been reversed. Images that were previously inhibitory (image \(0\) \(\to\) label "zero" \(\to\) inhibitory) are now excitatory (image \(0\) \(\to\) label "nine" \(\to\) excitatory), and vice versa. Swaps involve two randomly selected classes, which may have the same or opposing effects. 
\begin{figure}
  \centering
  \includegraphics[width=\textwidth]{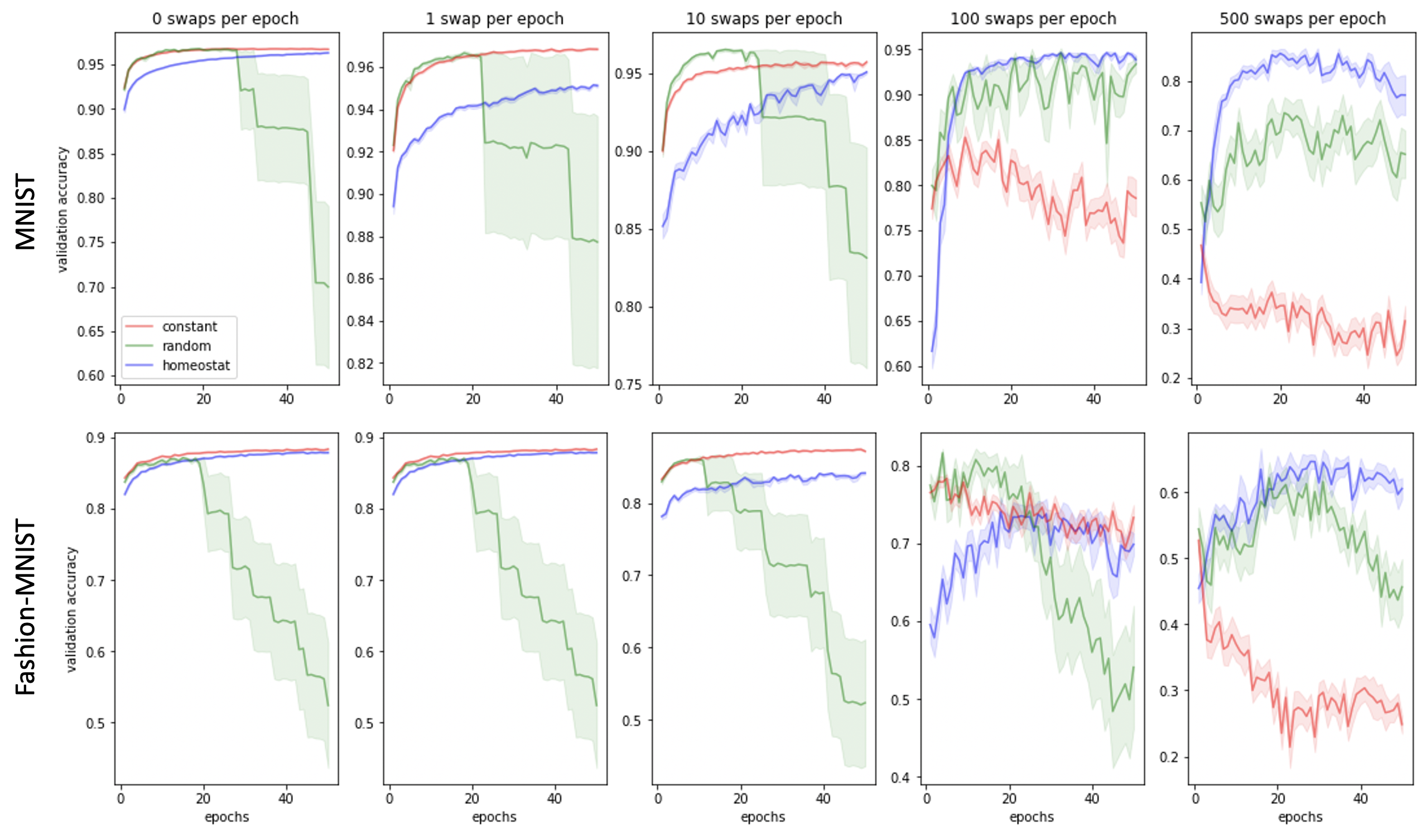}
  \caption{Homeostatic learners incur some performance penalty in environments with no or low concept shift, but are far superior under conditions of highest shift. Color-coded validation accuracies of learners with their learning rates homeostatically regulated (blue), randomly regulated (green), and held constant (red). Traces show mean +/- SEM over 20 replicates.
}
\end{figure}
\section{Experiments}
We compare homeostatic regulation of LR against two control conditions: a randomly regulated “wandering” learning rate, and a more conventional, constant learning rate. We characterize the conditions under which homeostatic regulation either imposes a performance penalty, or else allows a learner to smoothly adapt to changing conditions. All classification studies are performed with a multilayer perceptron with two hidden layers containing 80 and 60 units respectively, using the ELU activation function (Clevert et al 2015) and He initialization (He et al 2015). Minimum LR was clipped at \SI{1e-8}{}. We evaluate our method on two datasets, MNIST (Lecun et al 2010) and Fashion-MNIST (Xiao et al 2017). All experiments were originally performed in MATLAB. The source code has been ported to Python and is available on \href{https://github.com/kingsonman/homeostatic-neural-networks}{Github}.

\subsection{The homeostatic learner adapts to concept shift}
Testing across a wide range of rates of concept shift, measured in swaps per epoch of training, we find that in the stationary setting (no swapping) the conventional, constant-LR classifier is most accurate (Fig. 1, red traces in far left column). The homeostat (blue traces) nearly matches the constant-LR classifier's performance, which is remarkable because the homeostat has the seeming disadvantage of being vulnerable to its own mistakes. Illustrating how badly things could have gone, the randomly regulating LR classifier (green traces)  goes off the rails and shows large variance across replicates. The benefits of the homeostatic architecture become apparent at the highest intensities of concept shift (Fig. 1, right columns). At 500 swaps per epoch, the constant-LR classifier is overcome by change and falls to near chance level. The homeostat, on the other hand, is able to learn and even to improve despite extreme rates of concept shift.
\subsection{Homeostatic LR regulation is responsive to the prevailing rates of concept shift}
The homeostat tunes its learning rate to a level specific to the environment in which it finds itself (Supplementary Fig. 3). All learning rates are initialized at the same value (0.005) but the homeostat seeks a LR appropriate to the experienced rate of concept shift. It arrests its own rise in high shift environments and converges upon stable LR values. 
\subsection{The homeostatic learner adapts to second-order shifts, or "seasonality"}
We next created learning environments with seasonality, in which the rate of concept shift can vary over the course of training. We find that the homeostat maintains the most consistent performance across "calm" and "stormy" seasons, and rapidly recovers after the onset of a stormy period (Fig. 2). Note, however, that the homeostat shows an increase in variance and decrease in mean performance during the final "calm" period (right-most portions of second row, Fig. 2), which may indicate a memory effect from previous "stormy" seasons.
\begin{figure}
  \centering
  \includegraphics[width=.7\textwidth]{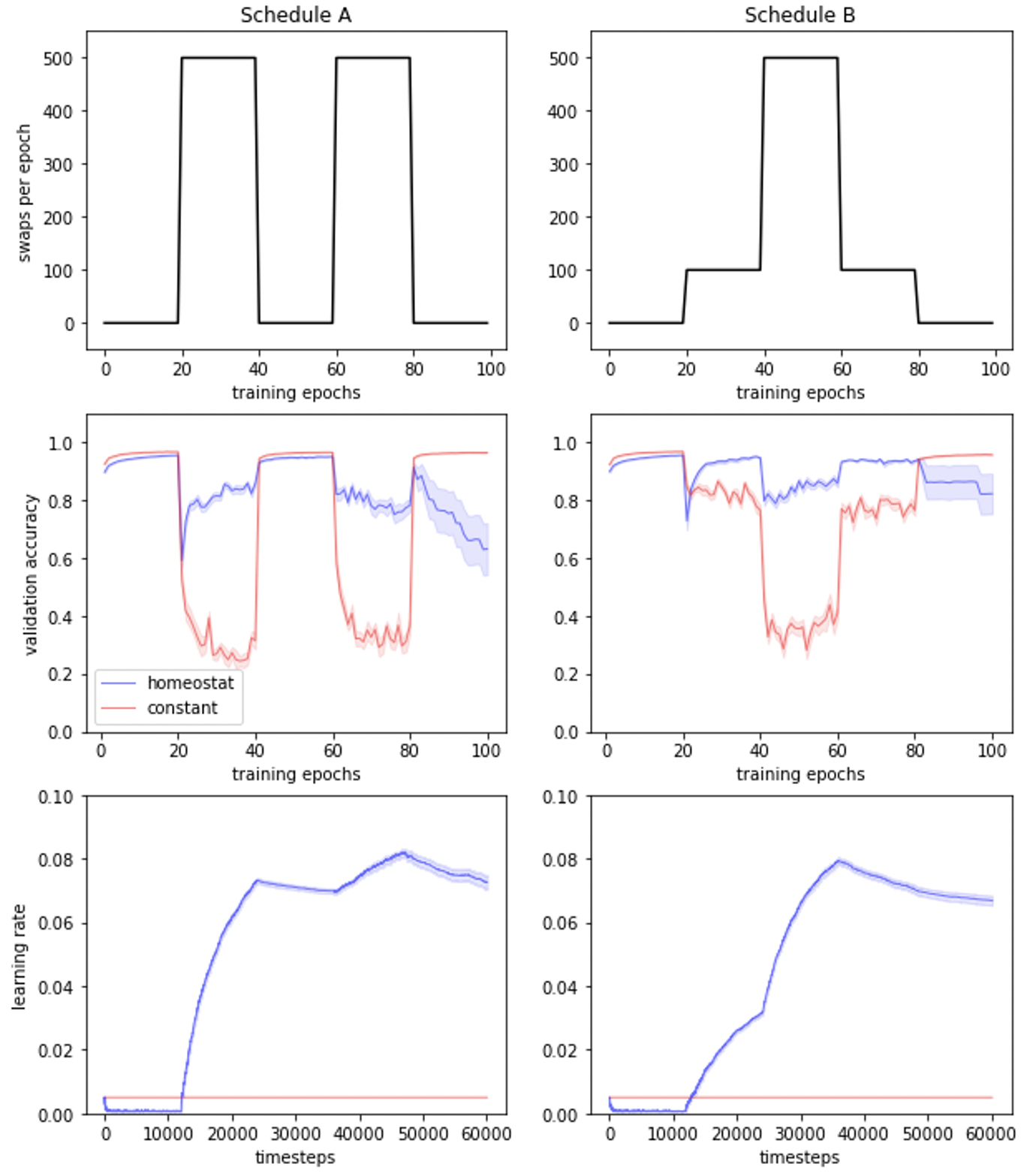}
  \caption{Accuracy and learning rate under “seasonality” of concept shift. Top row: Schedule A cycles between extreme rates of concept shift, while schedule B is more gradual. Middle row: The accuracy of the constant-LR classifier, in red, severely declines during stormy periods but returns to normal during calm periods. The homeostat, in blue, maintains good average performance across shifts in the rate of concept shift. Bottom row: The sequence of learning rates reveals that the homeostatic learner ratchets up its learning rate during stormy periods but is less inclined to reduce it during calm periods.}
\end{figure} 
\section{Discussion}
To summarize, we show that: 1) homeostatic learners are superior to conventional learners under concept shift, with the greatest advantage obtained under the greatest rates of shift; 2) homeostatic regulation imposes a slight performance penalty under static and low-shift environments; 3) homeostatic learners tune their learning rate in accordance with environmental conditions; and 4) homeostatic learners can adapt to second-order shift, or changes in the rate of environmental change. Although we find these converging results across the MNIST and Fashion-MNIST datasets we note that one possible limitation on the scope of our claims is the use of only these two datasets, each being somewhat limited in visual complexity and image size. 

Another possible limitation is the re-use of training data over many epochs, which limits the fundamental novelty of the concept shift. Although the labels and data are repeatedly shuffled, the classifier is never asked to learn from never-before-seen image patterns. In the real world, concept shift often co-occurs with some level of covariate shift. Not only do relationships change over time, but the predictors change as well. 

Although our method can dynamically adjust the learning rate, we did not benchmark it against LR optimizers such as ADAM (Kingma and Ba 2014) and other momentum-based methods. In the non-stationary setting the loss surface shifts over time and it is therefore inappropriate to accumulate previous gradients from an outdated loss surface. 

Finally, we are aware of the resemblance between reinforcement learning and our task of homeostatic self-regulation, though we argue that they should not be identified as the same. The objective here is not to maximize some arbitrary “reward” by massed trial-and-error. The object of the game is simply to keep playing the game. We specify a particular target to optimize: homeostatic well-being, crystallized as an internal parameter that controls the ongoing ability to make good decisions. For an excellent example of work in reinforcement learning that takes homeostatic logic into account see (Keramati and Gutkin 2014).

One way of explaining our homeostatic design is to say that it exposes an artificial neural network's thinking machinery to the consequences of its own "thoughts". A vulnerable learner with the meta-task of self-preservation is incentivized to better align with reality and to adapt to external change. The superior adaptability of the vulnerable learner illustrates the benefits of putting one’s own "skin in the game".

\begin{ack}

We gratefully acknowledge funding to KM and AD from Google and from the National Science Foundation under Grant No. IIS-2033961.

\end{ack}

\section*{References}

\medskip

{
\small
Clevert DA, Unterthiner T, Hochreiter S. Fast and accurate deep network learning by exponential linear units (elus). arXiv preprint arXiv:1511.07289. 2015.

Di Paolo EA. Homeostatic adaptation to inversion of the visual field and other sensorimotor disruptions. 2000.

He K, Zhang X, Ren S, Sun J. Delving deep into rectifiers: Surpassing human-level performance on imagenet classification. In: Proceedings of the IEEE international conference on computer vision 2015 (pp. 1026-1034).

Iizuka H, Di Paolo EA. Extended homeostatic adaptation: Improving the link between internal and behavioural stability. In: International Conference on Simulation of Adaptive Behavior 2008 Jul 7 (pp. 1-11).

Keramati M, Gutkin B. Homeostatic reinforcement learning for integrating reward collection and physiological stability. eLife. 2014;3:e04811.

Kingma DP, Ba J. Adam: A method for stochastic optimization. arXiv preprint arXiv:1412.6980. 2014.

LeCun Y, Cortes C, Burges CJC. The MNIST database of handwritten digits. http://yann.lecun.com/exdb/mnist/, 1994.

Man K, Damasio A. Homeostasis and soft robotics in the design of feeling machines. Nature Machine Intelligence. 2019;1(10):446-52.

Marder E, Goaillard JM. Variability, compensation and homeostasis in neuron and network function. Nature Reviews Neuroscience. 2006;7(7):563-74.

Moreno-Torres JG, Raeder T, Alaiz-Rodríguez R, Chawla NV, Herrera F. A unifying view on dataset shift in classification. Pattern Recognition. 2012;45(1):521-30.

Williams H, Noble J. Homeostatic plasticity improves signal propagation in continuous-time recurrent neural networks. Biosystems. 2007;87(2-3):252-9.

Xiao H, Rasul K, Vollgraf R. Fashion-mnist: a novel image dataset for benchmarking machine learning algorithms. arXiv preprint arXiv:1708.07747. 2017.


\appendix

\section{Supplementary Material}
\beginsupplement

\begin{figure}
  \centering
  \includegraphics[width=.8\textwidth]{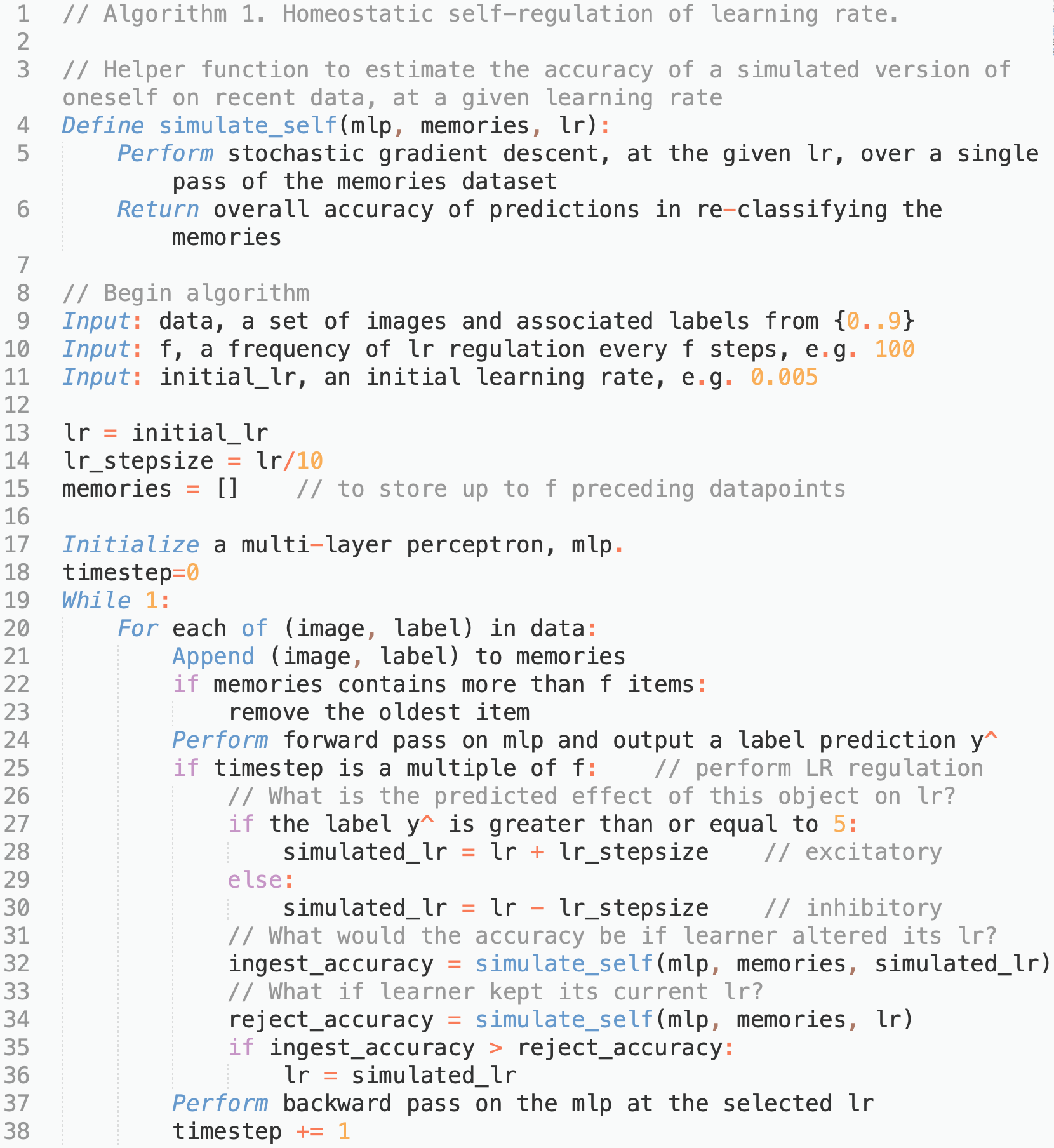}
  \caption{Algorithm: Pseudocode for homeostatic self-regulation of learning rate.}
\end{figure}

\begin{figure}
  \centering
  \includegraphics[width=.5\textwidth]{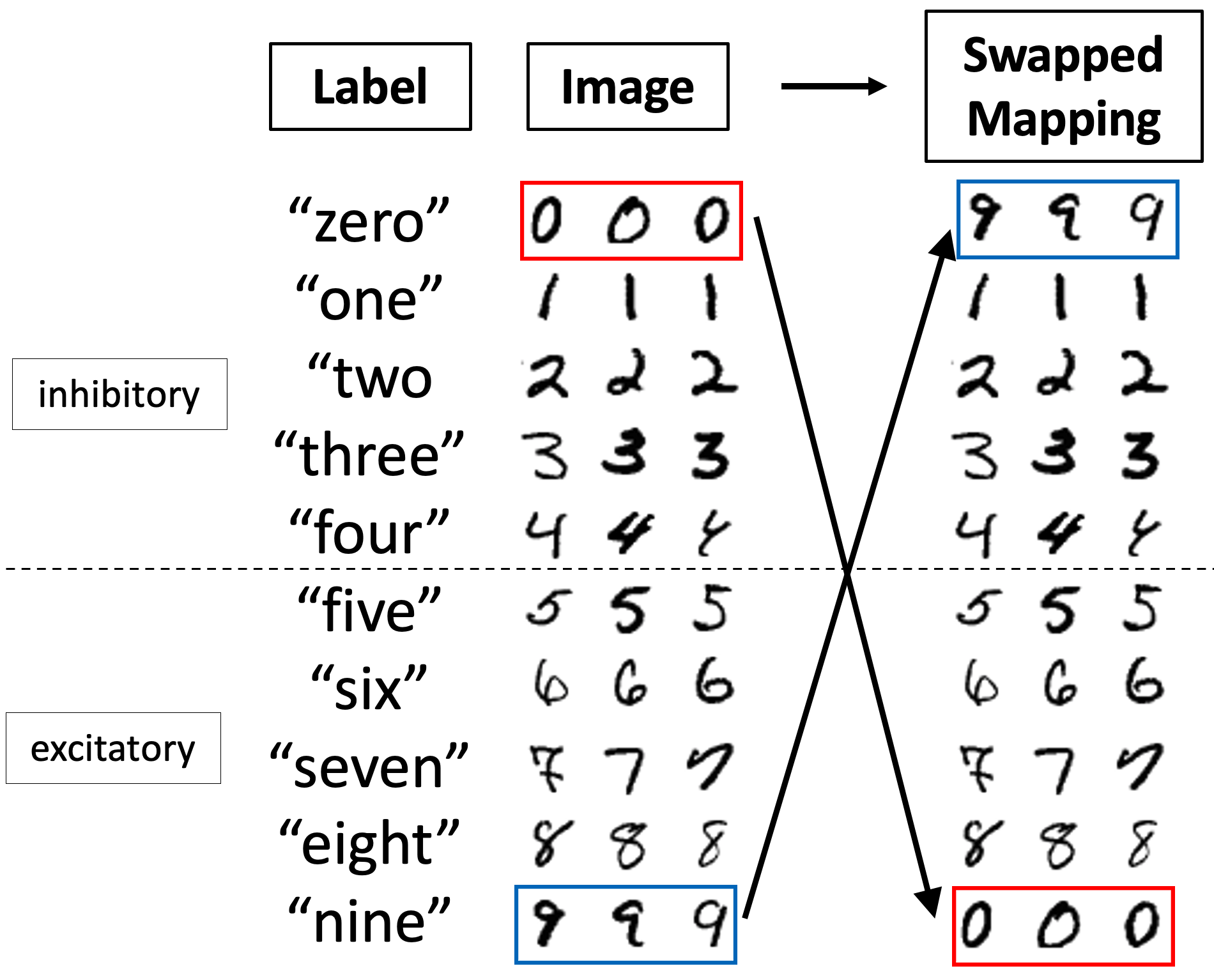}
  \caption{Concept shift is implemented by swapping the mapping between label and image between two randomly selected classes. Illustrated here on the MNIST dataset, the mappings for “zero” and “nine” are swapped. This swap will also invert the homeostatic effects expected for each number – a potentially hazardous situation for a vulnerable classifier.}
\end{figure}

\begin{figure}
  \centering
  \includegraphics[width=.6\textwidth]{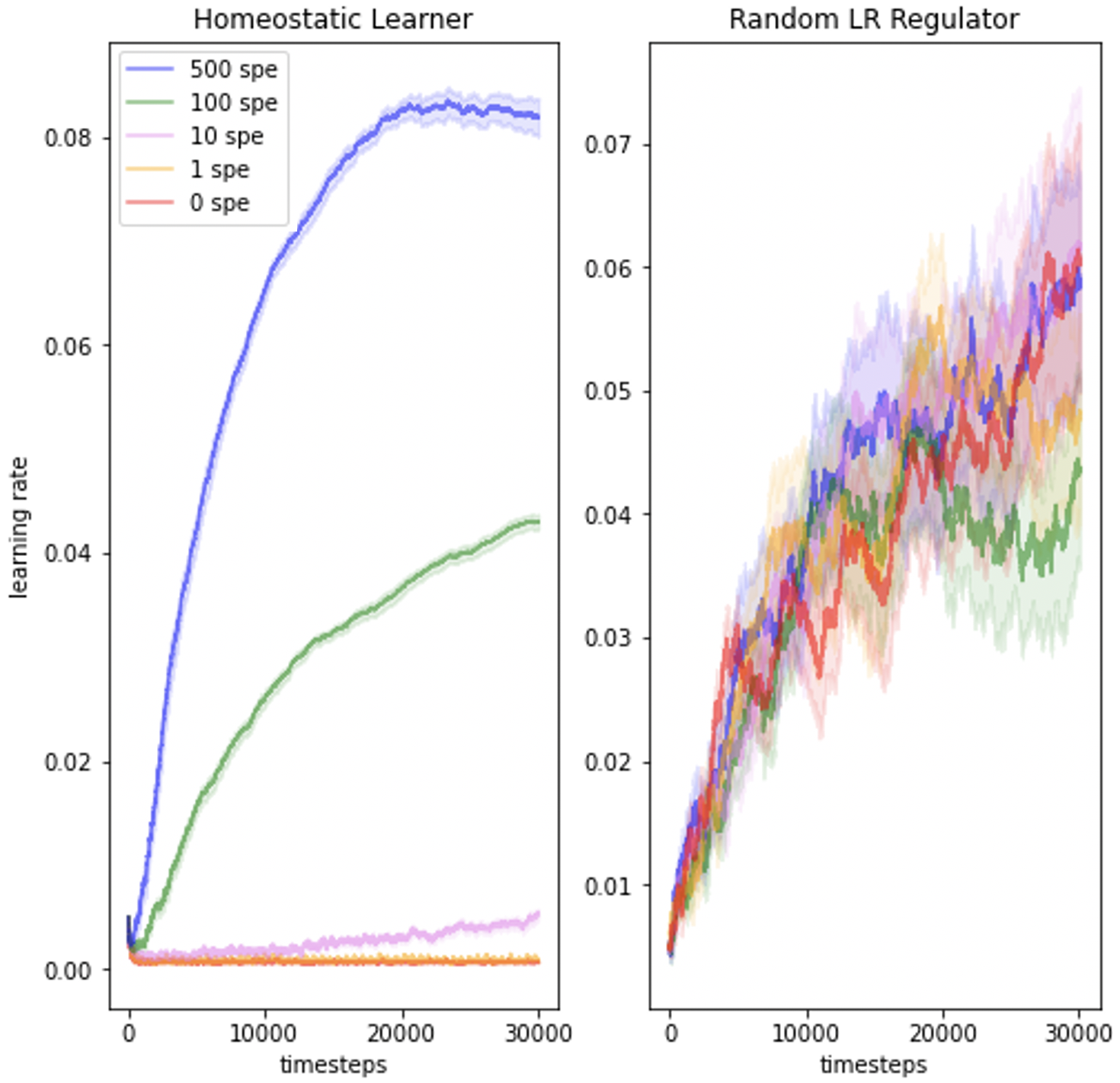}
  \caption{Learning rate sequencies of the two LR-regulating classifiers. The homeostatic learner seeks an LR appropriate to each level of concept shift, while the mean LR of the random regulator drifted upwards due to enforcement of a minimum non-zero LR. At 500 swaps per epoch (left, blue), the homeostat arrests its own LR growth and asymptotes. Data shown from MNIST only.}
\end{figure} 


\newpage
\section*{NeurIPS Paper Checklist}

\begin{enumerate}

\item {\bf Claims}
    \item[] Question: Do the main claims made in the abstract and introduction accurately reflect the paper's contributions and scope?
    \item[] Answer: \answerYes{} 
    \item[] Justification: We claim an exploration of homeostasis in artificial neural networks and deliver on such claims, showing benefits under certain situations. 

\item {\bf Limitations}
    \item[] Question: Does the paper discuss the limitations of the work performed by the authors?
    \item[] Answer: \answerYes{} 
    \item[] Justification: We discuss limits including dataset limitations, re-use of data for approximating concept shift, and intentional non-benchmarking against momentum-based optimizers. 

\item {\bf Theory Assumptions and Proofs}
    \item[] Question: For each theoretical result, does the paper provide the full set of assumptions and a complete (and correct) proof?
    \item[] Answer: \answerNA{} 
    \item[] Justification: We do not contribute a theoretical proof.

    \item {\bf Experimental Result Reproducibility}
    \item[] Question: Does the paper fully disclose all the information needed to reproduce the main experimental results of the paper to the extent that it affects the main claims and/or conclusions of the paper (regardless of whether the code and data are provided or not)?
    \item[] Answer: \answerYes{} 
    \item[] Justification: Code is available on Github to reproduce the results. Pseudocode is included in the appendix.

\item {\bf Open access to data and code}
    \item[] Question: Does the paper provide open access to the data and code, with sufficient instructions to faithfully reproduce the main experimental results, as described in supplemental material?
    \item[] Answer: \answerYes{} 
    \item[] Justification: See above. 

\item {\bf Experimental Setting/Details}
    \item[] Question: Does the paper specify all the training and test details (e.g., data splits, hyperparameters, how they were chosen, type of optimizer, etc.) necessary to understand the results?
    \item[] Answer: \answerYes{} 
    \item[] Justification: Experimental setting is explained to the extent possible in the page limits. Further details are in public code repository.

\item {\bf Experiment Statistical Significance}
    \item[] Question: Does the paper report error bars suitably and correctly defined or other appropriate information about the statistical significance of the experiments?
    \item[] Answer: \answerYes{} 
    \item[] Justification: We show mean +/- SEM over 20 replicates.

\item {\bf Experiments Compute Resources}
    \item[] Question: For each experiment, does the paper provide sufficient information on the computer resources (type of compute workers, memory, time of execution) needed to reproduce the experiments?
    \item[] Answer: \answerNo{} 
    \item[] Justification: The compute resources used were not remarkable; experiments were performed on a single entry-level consumer laptop. 
    
\item {\bf Code Of Ethics}
    \item[] Question: Does the research conducted in the paper conform, in every respect, with the NeurIPS Code of Ethics \url{https://neurips.cc/public/EthicsGuidelines}?
    \item[] Answer: \answerYes{} 
    \item[] Justification: Reviewed and confirmed.

\item {\bf Broader Impacts}
    \item[] Question: Does the paper discuss both potential positive societal impacts and negative societal impacts of the work performed?
    \item[] Answer: \answerNA{} 
    \item[] Justification: The author cannot conceive of any direct, realistic, and significant societal impacts of this foundational research. 
    
\item {\bf Safeguards}
    \item[] Question: Does the paper describe safeguards that have been put in place for responsible release of data or models that have a high risk for misuse (e.g., pretrained language models, image generators, or scraped datasets)?
    \item[] Answer: \answerNA{} 
    \item[] Justification: No such risks.

\item {\bf Licenses for existing assets}
    \item[] Question: Are the creators or original owners of assets (e.g., code, data, models), used in the paper, properly credited and are the license and terms of use explicitly mentioned and properly respected?
    \item[] Answer: \answerYes{} 
    \item[] Justification: MNIST and FMNIST are cited.

\item {\bf New Assets}
    \item[] Question: Are new assets introduced in the paper well documented and is the documentation provided alongside the assets?
    \item[] Answer: \answerNA{} 
    \item[] Justification: We do not release new assets/datasets.

\item {\bf Crowdsourcing and Research with Human Subjects}
    \item[] Question: For crowdsourcing experiments and research with human subjects, does the paper include the full text of instructions given to participants and screenshots, if applicable, as well as details about compensation (if any)? 
    \item[] Answer: \answerNA{} 
    \item[] Justification: No crowdsourcing.

\item {\bf Institutional Review Board (IRB) Approvals or Equivalent for Research with Human Subjects}
    \item[] Question: Does the paper describe potential risks incurred by study participants, whether such risks were disclosed to the subjects, and whether Institutional Review Board (IRB) approvals (or an equivalent approval/review based on the requirements of your country or institution) were obtained?
    \item[] Answer: \answerNA{} 
    \item[] Justification: No human subjects.

\end{enumerate}

\end{document}